\documentclass[letterpaper, 10 pt, conference]{ieeeconf}  

\usepackage[utf8]{inputenc} 
\usepackage[T1]{fontenc}    
\usepackage{url}            
\usepackage{booktabs}       
\usepackage{amsfonts}       
\usepackage{nicefrac}       
\usepackage{microtype}      
\usepackage{xcolor}         
\usepackage{todonotes}
\usepackage{caption}
\usepackage{wrapfig}

\usepackage{amsmath}
\usepackage{amssymb}
\usepackage{mathtools}
\usepackage{subcaption}

\usepackage{graphicx} 
\usepackage{placeins}
\usepackage{enumerate}
\usepackage{algorithmic}
\usepackage{algorithm}
\usepackage{float}

\makeatletter
\let\NAT@parse\undefined
\makeatother
\usepackage{hyperref}  

\title{Conditionally Combining Robot Skills using Large Language Models}

\makeatletter
\newcommand{\linebreakand}{%
  \end{@IEEEauthorhalign}
  \vspace{3mm}
  \hfill\mbox{}\par
  \mbox{}\hfill\begin{@IEEEauthorhalign}
}
\makeatother

\author{%
    K.R. Zentner\textsuperscript{*}\thanks{\textsuperscript{*}Continuation of work done as an intern at Google Brain}\\
    \textit{Univ. of Southern California}\\
    \texttt{kzentner@usc.edu}
  \and
    Ryan Julian\\
    \textit{Google DeepMind}\\
    \texttt{rjulian@google.com}
  \linebreakand
    Brian Ichter\\
    \textit{Google DeepMind}\\
    \texttt{ichter@google.com}
  \and
    Gaurav S. Sukhatme\textsuperscript{+}\thanks{\textsuperscript{+}GSS holds concurrent appointments as a Professor at USC and as an Amazon Scholar. This paper describes work performed at USC and is not associated with Amazon.}\\
    \textit{Univ. of Southern California}\\
    \texttt{gaurav@usc.edu}
}

\date{September 2023}

\begin{document}

\maketitle

\begin{abstract}
This paper combines two contributions.
First, we introduce an extension of the Meta-World benchmark, which we call ``Language-World,'' which allows a large language model to operate in a simulated robotic environment using semi-structured natural language queries and scripted skills described using natural language.
By using the same set of tasks as Meta-World, Language-World results can be easily compared to Meta-World results,
allowing for a point of comparison between recent methods using Large Language Models (LLMs) and those using Deep Reinforcement Learning.
Second, we introduce a method we call Plan Conditioned Behavioral Cloning (PCBC), that allows finetuning the behavior of high-level plans using end-to-end demonstrations.
Using Language-World, we show that PCBC is able to achieve strong performance in a variety of few-shot regimes, often achieving task generalization with as little as a single demonstration.
We have made Language-World available as open-source software at \href{https://github.com/krzentner/language-world/}{https://github.com/krzentner/language-world/}.
\end{abstract}


\section{Introduction}

Combining skills to perform new tasks is an area of active research in machine learning and robotics.
Skill reuse has several potential advantages, such as allowing a robot to efficiently re-use data from old tasks to rapidly learn new tasks.
Sequencing skills might also help address the difficulty of learning long-horizon tasks end-to-end.
Here, we propose one way of using text to combine skills that leverages the capabilities of recent large language models (LLMs).

Our approach uses a textual description of how to perform a task we term a ``conditional plan.''
A conditional plan lists a set of skills to use to perform a task and associates a linguistic condition under which each skill should be active.
See Table~\ref{sample-conditional-plan} for an example.
Because the plan expresses conditional behavior, our method is able to make use of an LLM without querying it while the robot is performing a task.
Instead, a much simpler neural architecture with several orders of magnitude fewer parameters implements the conditional behavior.
This maintains the runtime efficiency of other end-to-end approaches and allows fine-grained skills.
Because the conditional plan contains natural language, it can also improve interpretability and transparency by describing what skill the robot is using and why it is using it.
Further, it provides additional control compared to other approaches using LLMs, since a human can inspect or edit the conditional plan before it is used.

\begin{figure}[t]
    \begin{minipage}{\columnwidth}
        \includegraphics[width=\columnwidth]{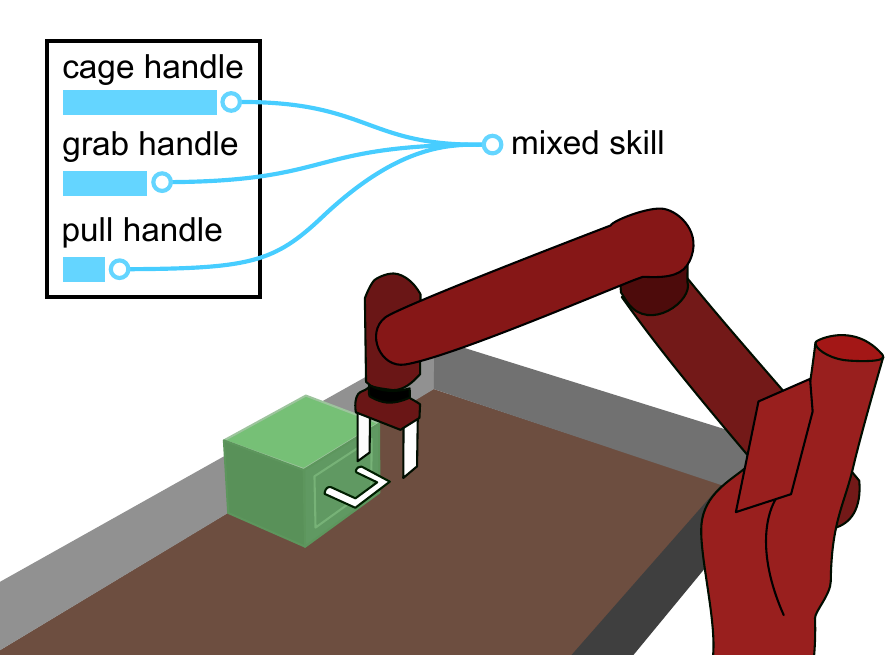}
        \vspace{-0.5cm}
        \caption{%
        This figure shows a simulated robot using our method.
        See Figure~\ref{fig:cond-plan-overview} for a detailed description of how it functions.
        }\label{fig:robot-example}
    \end{minipage}%
    \vspace{-5mm}
\end{figure}

In Section~\ref{language-world} we extend the popular ``Meta-World'' benchmark to create ``Language-World.''
Language-World includes a set of tools for working with language that allow us to evaluate our proposed method and compare it to several ablations, including using an LLM directly.

In Section~\ref{method}, we describe conditional plans and how we use an LLM to generate them.
Then, we propose a method called Plan Conditioned Behavioral Cloning to finetune the behavior of plans using end-to-end demonstrations.
Finally, we describe how to perform effective few-shot generalization using Cross-Task Co-Learning, including using PCBC.

The primary contributions of this work are as follows:

\begin{enumerate}
  \item A new benchmark, ``Language-World,'' which extends the Meta-World benchmark to make performing experiments with large language models on it practical.
  \item A method, Plan Conditioned Behavioral Cloning (PCBC), that allows finetuning the behavior of high-level plans using end-to-end demonstrations.
  \item Experimental demonstrations that PCBC and Co-Learning are able to achieve strong few-shot generalization results in ``Language-World.''
\end{enumerate}
\pagebreak

\section{Related Work}\label{related-work}

\paragraph{Large Language Models}
Recent work in language modeling has resulted in ``large language models'' (LLMs), which contain billions of neural network parameters and demonstrate powerful zero and few-shot reasoning capabilities.
In this work, we experiment with using three such LLMs: GPT-3 ~\cite{brown2020language}, GPT-3.5~\cite{gpt35}, and PaLM 2~\cite{Chowdhery2022PaLMSL, palm2}.
This is an area of active research, and additional LLMs have become available since we began this work, including GPT-4~\cite{openai2023gpt4}, LLaMa~\cite{touvron2023llama}, and Claude~\cite{bai2022training}.
Several methods to improve the utility of LLM output for downstream tasks have also been proposed, including finetuning with RL \cite{ouyang2022training}, finetuning with supervised learning \cite{rafailov2023direct} or improved prompting \cite{Wei2022ChainOT, Wang2022SelfConsistencyIC, Zhou2022LeasttoMostPE}.
In this work, we experiment with a variety of prompt formats that make use of chain-of-thought \cite{Wei2022ChainOT}, which is often able to improve the quality of LLM output with minimal effort.

\paragraph{Deep Reinforcement Learning (RL), End-to-End (E2E) Learning for Robotics}
Learning robotic capabilities via RL has been studied for decades~\cite{kober2013reinforcement,mahadevan1992automatic,lin1992reinforcement,smart2002effective}.
More recent advances in neural networks that allow feature learning and fitting to complex high-dimensional functions have allowed end-to-end training of neural policies using RL~\cite{franccois2018introduction,mnih2013playing} and imitation learning (IL)~\cite{Zhang2017DeepIL,Codevilla2017EndtoEndDV}.
A number of simulated environments for benchmarking these end-to-end methods on robotic tasks exist, including Meta-World~\cite{yu2019meta}, which we extend in this paper.

\paragraph{Skills, Options, and Hierarchy in E2E Learning}
Learning reusable skill libraries is a classic approach~\cite{gullapalli1994acquiring} for efficient acquisition and transfer of robot motion policies.
Prior to the popularity of E2E methods, several methods \cite{pastor2012asm,rueckert2015movprim,zhou2020incremental} were proposed for acquiring a set of reusable skills for robotic manipulation.
More recent E2E methods have been proposed for learning manipulation skills
\cite{yang2020multi,wulfmeier2021data,kroemer2015towards}
as well as skill decomposition methods for learning and adaptation in locomotion and whole-body humanoid control~\cite{peng2019mcp,hasenclever2020complementary,merel2020catch,li2020learning,hausman2018learning,julian2018scaling}.
Although these methods have demonstrated some improvements to the sample efficiency of RL, significant improvements in complex environments remain elusive.

\paragraph{Large Language Models as Agents}
Several recent works attempt to produce agents with useful zero-shot behavior by leveraging the generalization capabilities of large language models while mitigating their weaknesses at multi-step reasoning.
Some approaches, such as \cite{saycan2022arxiv,singh2023progprompt,Wang2023DescribeEP}, use an LLM to choose from a set of high-level actions described with natural language.
Other approaches, such as \cite{Liang2022CodeAP,Qin2023ToolLW}, use an LLM to generate code which is then executed to produce behavior.
The method proposed in this paper exists in a middle ground between these approaches, where an LLM is used to generate code in a particular format that allows actions to be described with natural language, and the behavior of the program (i.e. conditional plan) can be tuned E2E.





\section{Language-World}\label{language-world}

\begin{table*}[t]
    \centering
    \vspace{3mm}
    \setlength\tabcolsep{0pt}
    \begin{tabular*}{0.8\linewidth}{@{\extracolsep{\fill}} ll }
    Condition & Skill \\
    \midrule
    the gripper is closed and not near the drawer handle & open the gripper  \\
    the gripper is not near the drawer handle & move the gripper above the drawer handle  \\
    the gripper above the drawer handle & move the gripper down around the drawer handle \\
    the gripper is open and around the drawer & close the gripper \\
    the gripper is closed and around the drawer & pull the drawer open \\
    \bottomrule
  \end{tabular*}
  \caption{%
  Conditional Plan for \texttt{drawer-open}
  }\label{sample-conditional-plan}
    \vspace{-2mm}
\end{table*}

The Meta-World benchmark has emerged as a popular simulated environment for evaluating machine learning methods in the context of robotics, receiving over 150 citations in the past year alone.
It provides 50 robotic manipulation tasks with a continuous range of random goals available for each task.
The last year has also seen a rapid increase in interest with using large language models (LLMs) for robotics.
Ideally, research using LLMs for robotics would use benchmarks that allow quantitative comparisons to methods that do not use LLMs.
Language-World makes it relatively easy to perform these comparisons by providing three items that are useful, or necessary when using LLMs with Meta-World.

First, Language-World provides a brief natural-language description of each task e.g. ``push the button on the coffee machine.''
Second, Language-World provides a query answering function (QAF), which allows the evaluation of a semi-structured language query about a Meta-World state.
This module performs similarly to a VQA model optimized for Meta-World, while avoiding the overhead of rendering images.
By using the query answering function, methods that use language but do not deeply integrate visual processing with language can be efficiently evaluated on Meta-World tasks.
The third item Language-World provides is a set of 30 scripted skills.
These scripted skills can be used to reliably perform all tasks in the MT10 task set but can be applied to any of the 50 tasks.
However, note that the method we propose in this work does not use the scripted skills.
These three items allow Language-World to be used to perform simulated robotics experiments that can be easily compared to existing results in the literature.

\paragraph{\textbf{Task Descriptions}}\label{task-descriptions}
Each of the 50 tasks in MT50-language has a one-sentence natural language description of the task.
These descriptions can be used in a variety of ways, such as for conditioning a multi-task policy, or as inputs to an LLM.
In our experiments below, we use these descriptions as conditioning in a baseline method (Descriptor Conditioned BC), as well as to prompt an LLM to generate a plan in Plan Conditioned BC.

\paragraph{\textbf{Query Answering Function}}\label{query-answering-function}
The Language-World query answering function (QAF) is able to evaluate semi-structured textual queries on a Meta-World state.
These queries resemble natural language e.g. ``the gripper is around the puck.''
The QAF can evaluate 13 simple geometric relationships between all objects in the task, such as ``near,'' ``left of,'' ``in front of,'' or ``below.''
The QAF can evaluate these relationships between all objects in all 50 tasks, including the robot's gripper as well as objects not present in the observation but at fixed locations, such as walls.
It also can evaluate other useful cases, such as whether an object is touching the table, or if the robot's gripper is closed.
Finally, this function can handle negation, simple conjunctions of the above, and perform basic inference to identify repeated subjects, allowing evaluating conditions like ``Is the gripper open and not above the puck''.
The QAF also provides a list of all supported queries, so queries outside of the supported set can be mapped to the nearest supported query using a sentence embedding or using string edit distance.
For simplicity of interpretation in this work, we use string edit distance when necessary.

\begin{figure}
    \vspace{-5mm}
    \begin{minipage}{0.48\textwidth}
        \includegraphics[width=1.1\textwidth]{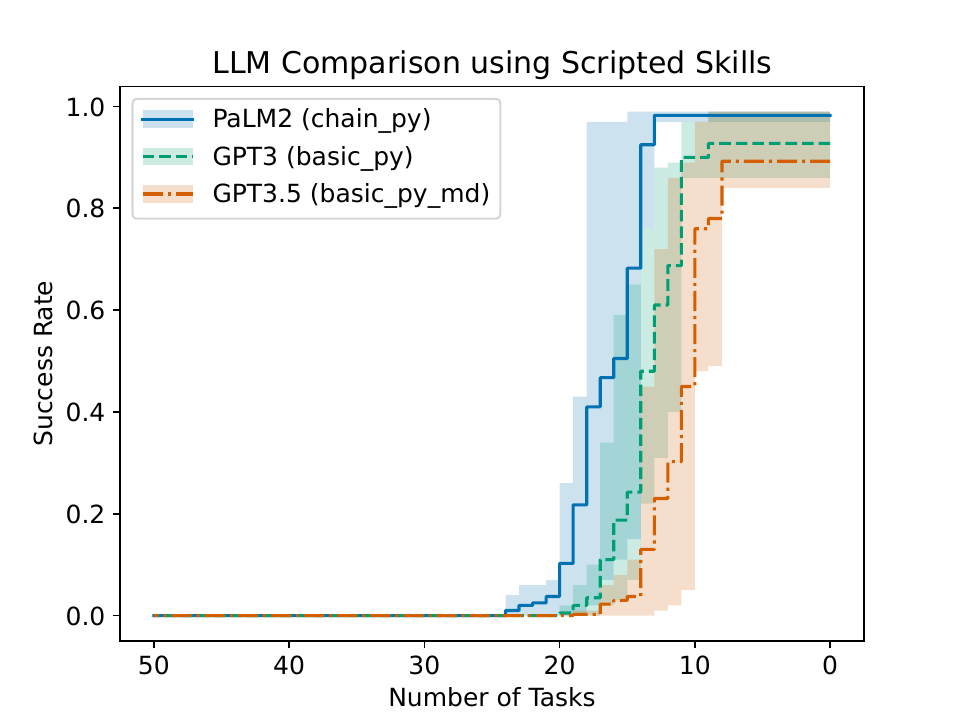}
        \vspace{-0.5cm}
        \caption{%
        This figure shows the performance of different LLM's on MT50-language using the scripted skills from MT10-language as a cumulative distribution function over tasks.
        Conditional plans were evaluated using the method described in Section~\ref{language-world-llm-evaluation}, and this figure shows the range of performance across 4 plans per task for each LLM using the plan format that performed best with that LLM.
        The LLMs are able to generalize to 5-10 additional tasks outside of MT10-language, despite using only scripted skills.
        }\label{fig:scripted-skills-overall}
    \end{minipage}
    \vspace{-7mm}
\end{figure}

\paragraph{\textbf{Scripted Skills}}\label{scripted-skills}
Language-World provides 30 scripted skills, each of which has a brief natural language description (e.g. ``put the gripper above the drawer handle'').
These scripted skills have been tested using a hand-crafted mapping between queries and skills, and are able to perform all of the tasks in MT10-language with a success rate of over 90\%.
These scripted skills are stateless linear controllers extracted from the Meta-World scripted policies.
Because most tasks require use of more than one skill, there are more skills than tasks, even though some tasks share skills.
We use these scripted skills to compare the performance of 3 LLMs in Figure~\ref{fig:scripted-skills-overall} and in evaluating different plan formats in Figure~\ref{fig:scripted-skills-plan-format-comparison}.
However, we \textbf{do not use scripted skills in our remaining experiments.}

\paragraph{\textbf{Task Formalism}}\label{task-formalism}
Language-World uses the description of a task that is frequently used when performing multi-task RL with Meta-World.
Specifically, each of the 50 available named tasks $\tau$ is an infinite horizon fully observable Markov Decision Process with a non-Markovian indicator function that measures success on a given episode.
The episodes of $\tau$ \textit{must} be sampled using 500 sequential states, beginning with a random initial state drawn from a continuous uniform distribution defined by the benchmark, as well as a randomized goal (this configuration is sometimes referred to as MT10-rand or MT50-rand in the literature).
We refer to using the tasks from MT10 augmented with language as MT10-language, and equivalently with MT50 and MT50-language. Note that MT10 is a subset of MT50.

\section{Method}\label{method}

In this section, we propose a method for controlling a robot that makes use of language.
We will later evaluate this method using two of the tools described above in Language-World: the \hyperref[task-descriptions]{task descriptions} and \hyperref[query-answering-function]{query answering function (QAF)}.

\textbf{Core Idea:} The core idea of our method is to query the large language model (LLM) once per task to produce a fixed mapping of queries to skills (a ``conditional plan'').
Then, at each state, we will use a query evaluation module to evaluate each of those queries and perform a skill that corresponds to the true queries.
By designing a neural architecture that interprets conditional plans in an end-to-end differentiable way, we are able to finetune the behavior of the generated plans from demonstrations, which we call plan conditioned behavioral cloning (PCBC) and describe in detail in Section~\ref{cond-plan-architecture}.

Besides the strong experimental results we present in Section~\ref{cond-plan-experiments-lw}, plan conditioning has several promising conceptual aspects.
By its design, plan conditioning splits the end-to-end problem into three stages, while preserving the ability for end-to-end training.
The first stage, plan generation (which we describe in Section~\ref{cond-plan-generation}), corresponds approximately to ``task planning,'' and can be performed before a task is attempted, allowing oversight or input from a human operator.
The second stage, query evaluation, permits significant implementation flexibility.
Query answering could be performed by a visual question answer (VQA) model or using value functions (as in \cite{saycan2022arxiv}) and finetuned as part of end-to-end training.
Alternatively, query answering could use any of a variety of perception methods that have been well-studied in the robotics literature.
As our experiments show, constraining the generated queries to those that the query-answering module is engineered to perform can be highly effective.
This allows leveraging ``internet scale'' models without discarding the significant progress made in robotics perception methods.
The third stage, action decoding, could also be performed by a variety of methods, allowing for much higher control frequencies than the query answering module.







\pagebreak


\subsection{Conditional Plan Generation}\label{cond-plan-generation}
In order to use a conditional plan to solve a task, we first must generate that plan.
Formally, we consider a conditional plan $P_\tau$ that describes how to perform some task $\tau$ to be a set of (condition, skill) tuples $(c_i, k_i)$, where
each $k_i$ is a natural language description of a skill, which we will embed into a continuous latent space.
Depending on the benchmark $c_i$ is either a semi-structured condition (in Language-World), or a natural language condition that can be evaluated by a visual question answering (VQA) model.
Because LLMs only generate text, in order to use them to generate a conditional plan we need a \textit{plan format} that will allow encoding and decoding conditional plans from text.
We experiment with nine different plan formats, and found that different plan formats perform best for different LLMs.
For example, GPT-3.5, which has been finetuned on markdown format text, performs best with \texttt{basic\_py\_md}.

To produce a prompt given a plan format, we first manually wrote plans for each of the tasks in MT10-language.
Then, we generated a prompt for each task in MT50-language by taking manually written plans from three other tasks, and formatting them using the plan format.
We chose the three other tasks by using \texttt{pick-place} (because it contains a set of skills consistently useful across many tasks), as well as the two other tasks which had task descriptions with the smallest string edit distance from the description of the task we were prompting the LLM to generate a plan for.
When prompting, we never provided a plan for the task we were currently prompting for, to avoid the LLM repeating back the manually written plan.
We end the prompt with the name of the task as well as the task description, which is also encoded using the plan format.
We prompted each LLM four times for each combination of task and plan format.

\subsection{LLM Experiments}\label{language-world-llm-evaluation}
\begin{figure}
    \begin{minipage}{0.48\textwidth}
        \vspace{-7mm}
        \includegraphics[width=1.1\textwidth]{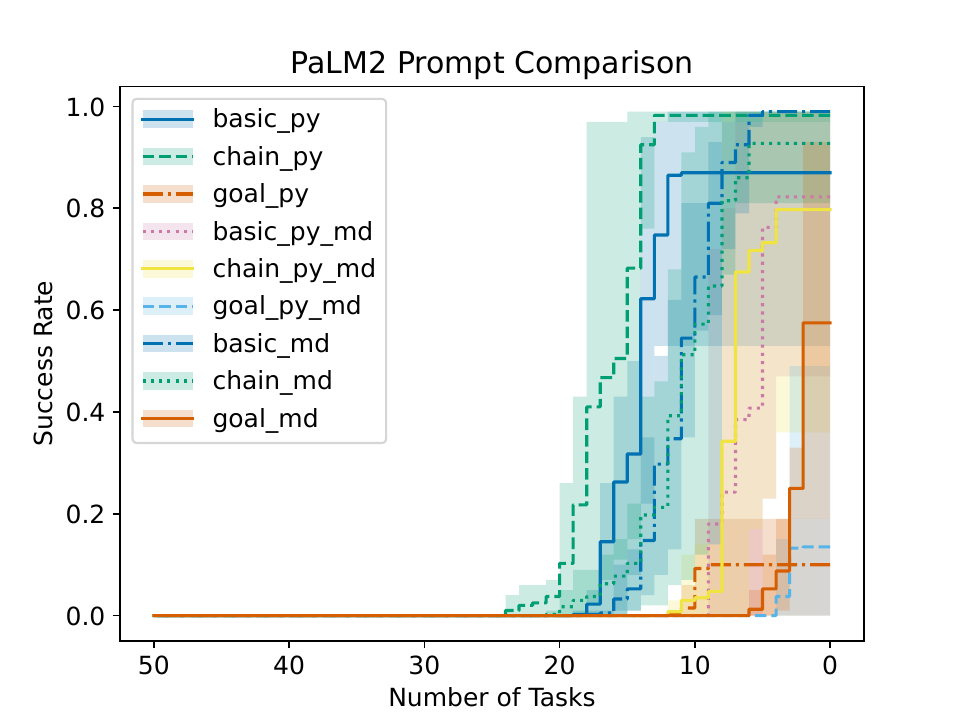}
        \vspace{-5mm}
        \caption{%
        This figure shows the performance of PaLM 2 using different plan formats on MT50-language using the scripted skills from MT10-language as a cumulative distribution function over tasks.
        Plan formats have a significant effect on performance, varying the LLM from being able to barely perform 3 tasks to being able to reliably perform 15 tasks.
        Shaded region is between minimum and maximum performance across 4 plans produced by the LLM for each task.
        }\label{fig:scripted-skills-plan-format-comparison}
    \end{minipage}
\end{figure}

In order to efficiently evaluate the performance of different combinations of LLM and plan format, we ran each generated plan using the Language-World \hyperref[query-answering-function]{query answering function (QAF)} and \hyperref[scripted-skills]{scripted skills}.
Because the QAF only supports a finite set of (several hundred) semi-structured natural language queries, and the scripted skills consist of only 30 skills, this required us to map each condition $c_i$ to the nearest query $q_i$ supported by the QAF, and each natural language skill $k_i$ to the scripted skill with the nearest description.
We experimented with using both distance in an embedding space as well as using string edit distance.
Because all LLMs fairly consistently matched the expected format, we found that edit distance performed sufficiently well, and used that to perform this mapping.
The best results for each LLM using these plans are show in Figure~\ref{fig:scripted-skills-overall}.
In Figure~\ref{fig:scripted-skills-plan-format-comparison} we compare different plan formats using PaLM 2, showing the importance of careful prompting to achieve the best results.

\begin{figure}
    \begin{minipage}{\columnwidth}
        \includegraphics[width=\columnwidth]{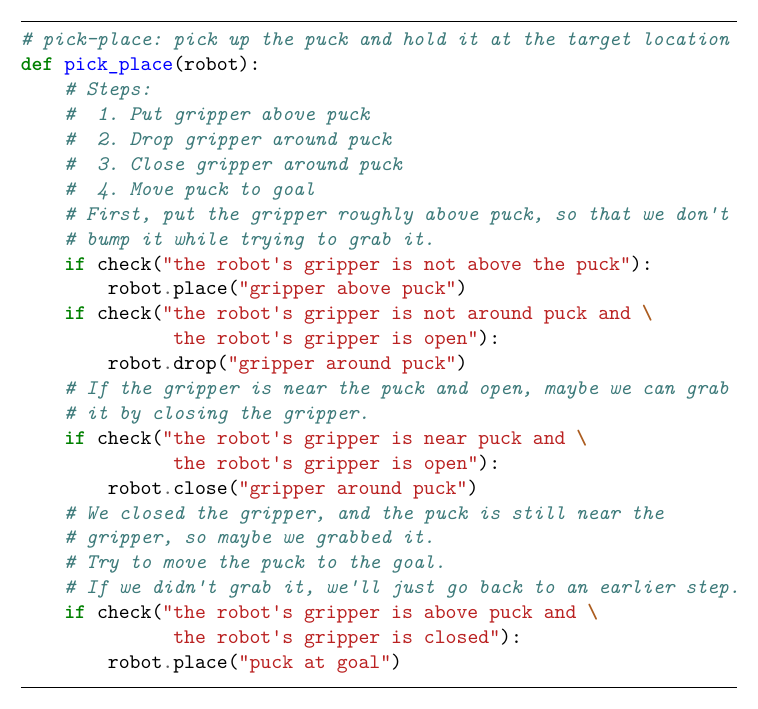}
        \vspace{-0.5cm}
        \caption{%
        An example of the \texttt{pick-place} plan in the \texttt{chain\_py} format. Although formatted as code, we do not evaluate this code directly.
        In Section~\ref{cond-plan-architecture} we describe how to evaluate this code using PCBC, which allows finetuning the behavior of this program using end-to-end demonstrations.
        Our method extracts the condition in each if statement, and transform each function call into a skill description by turning the code into equivalent natural language.
        For example, \texttt{robot.place("gripper above puck")} becomes the skill description ``place the gripper above the puck,'' via a simple regex search and replace.
        }\label{fig:prompt-format-example}
    \end{minipage}%
\end{figure}

We call the best plan format we found \texttt{chain\_py}, which uses a chain-of-thought \cite{Wei2022ChainOT} style prompt, with conditions and skills both encoded in a format that appears similar to python code.
An example of a plan in this format can be seen in Figure~\ref{fig:prompt-format-example}.

\pagebreak
\subsection{Plan Conditioning}\label{cond-plan-architecture}
\begin{figure}
    \begin{minipage}{\columnwidth}
        \vspace{2mm}
        \includegraphics[width=\columnwidth]{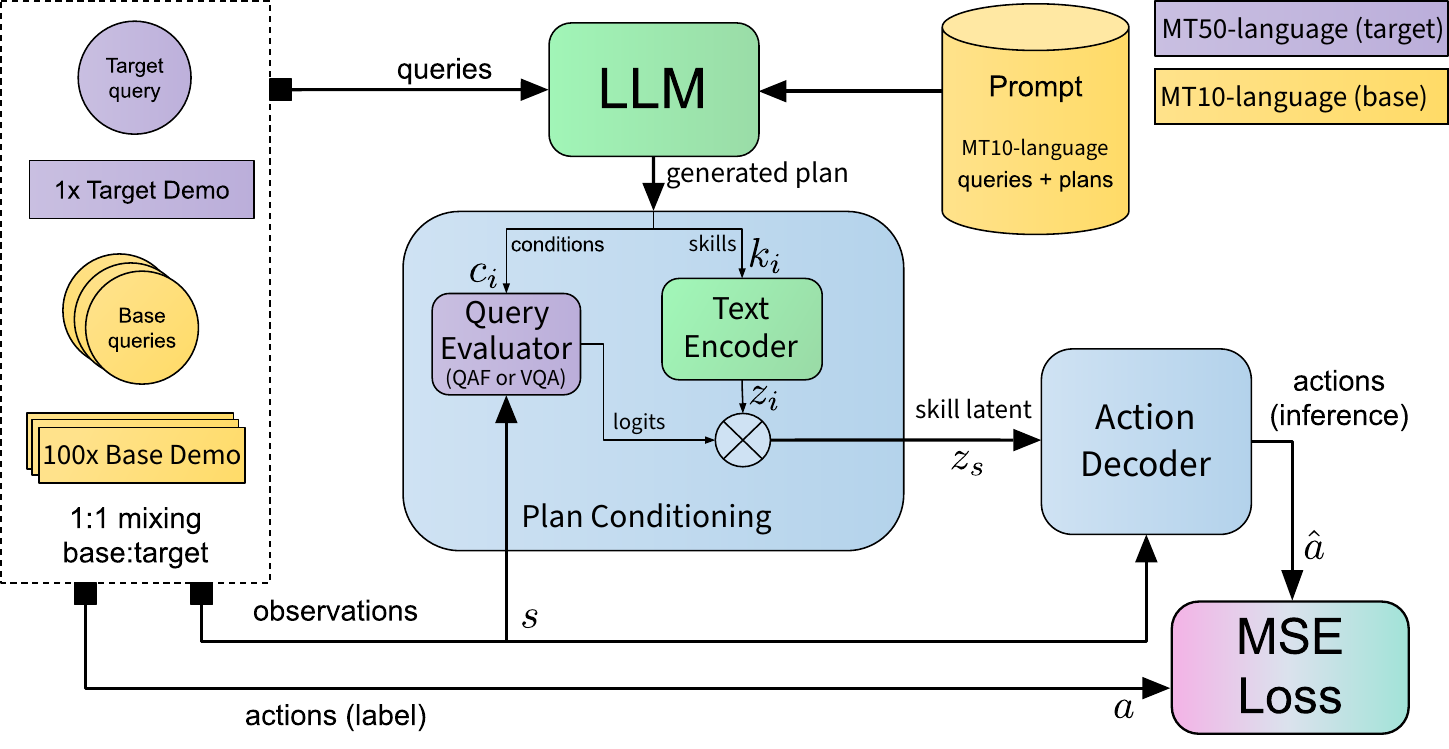}%
        \vspace{-2mm}
        \caption{%
        This figure shows our proposed neural architecture and training setup on Language-World.
        The data setup for one-shot training is shown on the left.
        PCBC finetunes the Action Decoder to match demonstrations using the MSE Loss, and produces gradients that could be used to tune the QAF.
        }\label{fig:cond-plan-overview}
    \end{minipage}%
    \vspace{-6mm}
\end{figure}

Running a conditional plan with a set of scripted skills does produce some amount of zero-shot generalization.
However, we would also like to be able to combine the advantages of conditional plans with end-to-end machine learning\textemdash without requiring scripting skills.
To that end, we define an end-to-end differentiable neural architecture that interprets a conditional plan as follows.
To perform a task $\tau$ using a conditional plan $P_\tau$, first we encode each skill description $k_i$ into a \textit{skill latent} vector $z_i$.
Then, to select an action on a particular state $s$, we evaluate each condition $c_i$, to form an attention vector across each  skill latent $z_i$, and use the softmax of that attention vector to mix the skill latents into a single skill latent $z_s$.
That skill latent $z_s$ is then provided with the current state $s$ to an action decoder, which produces an action for the state.
The skill encoding and action decoder are trained using data from multiple tasks, as described in Section~\ref{data-mixing}.
See Figure~\ref{fig:cond-plan-overview} for a visual depiction of this process.

\subsection{Conditional Plan Experiments}\label{cond-plan-experiments-lw}

One of the most promising aspects of LLMs is their \mbox{few-shot} generalization capabilities.
In this section we present experiments that demonstrate PCBC's ability to extend this few-shot generalization into the robotics domain, and compare against an architecture that produces actions conditioned on only the task description and state, which we call descriptor conditioning (DC).
These experiments use the best of four plans generated with the best (plan format, LLM) combination found in Section~\ref{language-world-llm-evaluation}.
We run both neural architectures in three different data configurations shown in Table~\ref{data-requirements}.

\vspace{-1mm}
\begin{table}[H]
    \centering
    \setlength\tabcolsep{0pt}
    \begin{tabular*}{\linewidth}{@{\extracolsep{\fill}} lllll }
    Configuration & MT10 Demos. & MT50 Demos. & \#Models & Scripted Skills \\
    \midrule
    scripted skills & 0 & 0 & 0 & Yes \\
    zero-shot & 100 per task & 0 & 1 & No \\
    few-shot & 0 & 10 per task & 1 & No \\
    one-shot & 100 per task & 1 per task & 50 & No \\
    \bottomrule
  \end{tabular*}
  \caption{%
  Data used in Figures~\ref{fig:zero-shot-mt50}~and~\ref{fig:one-shot-mt50}.
  We used the same data pipeline, loss function, and optimizer for PCBC and DC.
  In all cases this is significantly less data than typically used by RL algorithms, which often require at least 10,000 episodes per task to achieve a non-zero success rate.
  \vspace{-1mm}
  }\label{data-requirements}
\end{table}

\begin{figure}[H]
    \centering
        \vspace{-3mm}
        \includegraphics[width=1.1\linewidth]{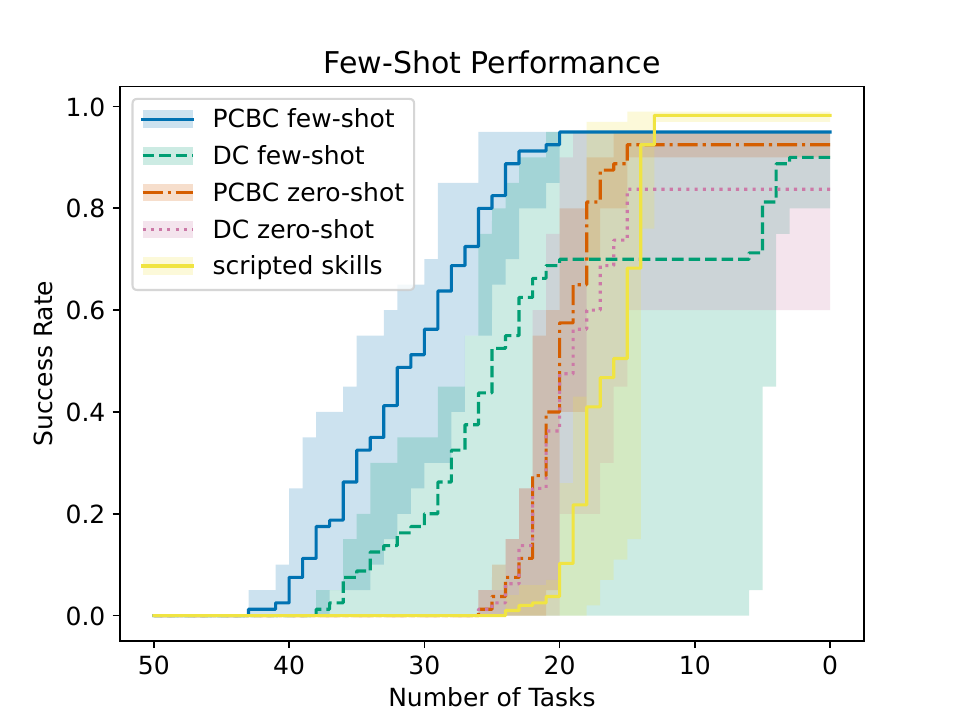}
        \vspace{-5mm}
        \caption{%
        This figure shows both few-shot and zero-shot performance of plan conditioned behavioral cloning (PCBC) as well as descriptor conditioning (DC) and scripted skills on MT50-language.
        Runs marked zero-shot were pre-trained from 100 demonstrations of MT10-language and were only provided a task description for MT50-language.
        Runs marked few-shot received 10 demonstrations for each task in MT50-language, as well as a task description.
        In both cases, one ``universal'' policy is learned for all tasks.
        End-to-end training improves over scripted skills, and plan conditioning (PCBC) maintains a higher consistent level of performance across many tasks than descriptor conditioning (DC).
        Shaded region is between min and max performance across 4 seeds.
        }\label{fig:zero-shot-mt50}
        \vspace{-5mm}
\end{figure}

\begin{figure}[H]
    \begin{minipage}{0.48\textwidth}
        \vspace{-10mm}
        \includegraphics[width=1.1\textwidth]{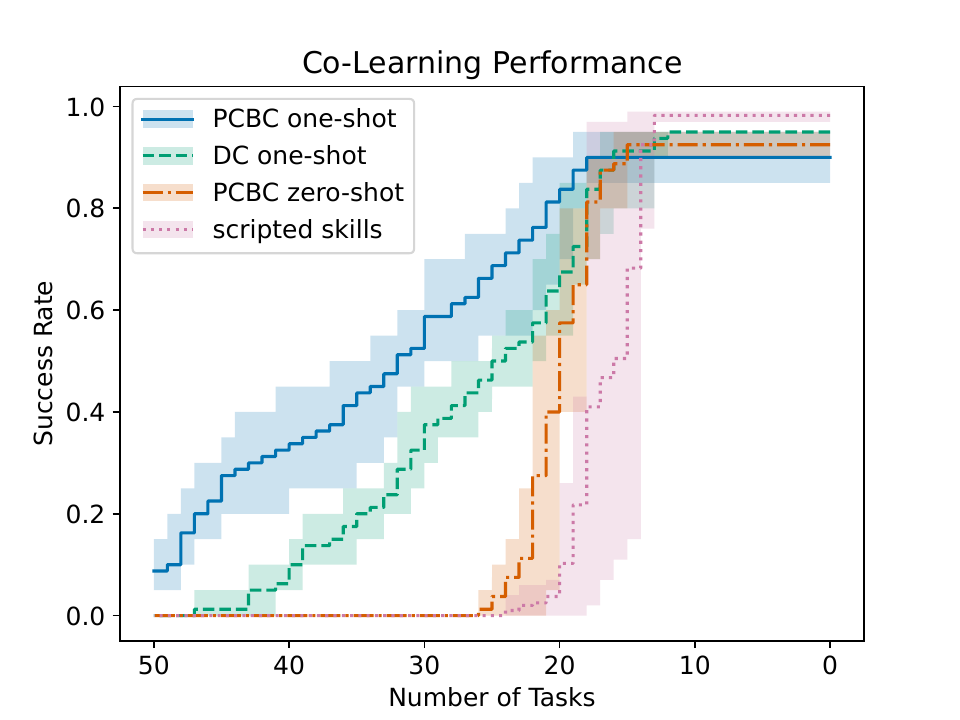}
        \caption{%
          This figure shows how adding a single demonstration to the zero-shot setting described in Figure~\ref{fig:zero-shot-mt50} results in a significant increase in performance across several tasks, and non-zero performance on every task, despite each task having randomized goal locations and initial states.
          The single demonstration of the MT50-language task is combined with 100 demonstrations of each MT10-language task using the co-learning method described in Section~\ref{data-mixing} to train a single model for each task.
          Shaded region is between minimum and maximum performance across 4 seeds.
        }\label{fig:one-shot-mt50}
    \end{minipage}
\end{figure}

\subsection{Cross-Task Co-Learning via 1:1 Data Mixing}\label{data-mixing}
\begin{figure}[H]
    \begin{minipage}{\linewidth}
        \includegraphics[width=\textwidth]{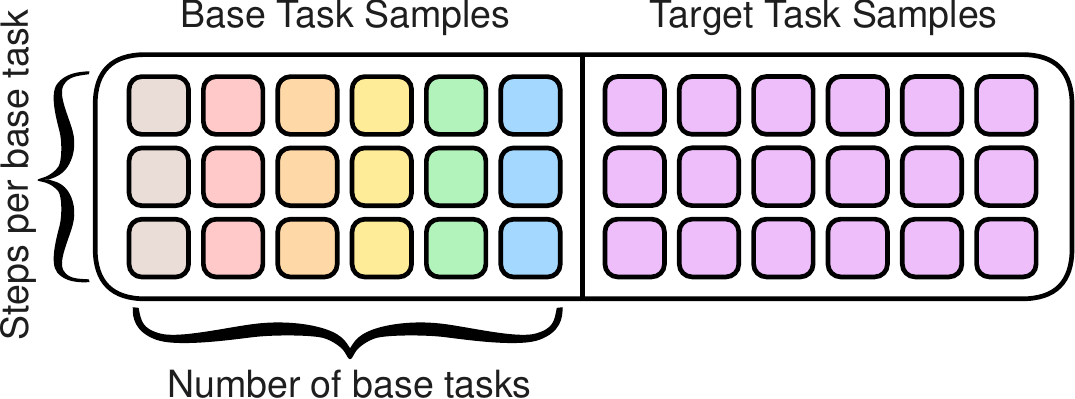}
        \caption{%
            A co-learning minibatch used in one-shot training.
            Each cell contains a (task, state, action) tuple which is used with the end-to-end BC loss to optimize the policy.
            Because there are significantly more target task tuples than tuples for any particular base task, the model will primarily be optimized for the target task while being regularized by the base task demonstrations.
            This regularization allows the trained policy to be robust to randomizations in the initial and goal state of a task, despite being trained on only a single demonstration with only a single initial state and goal state, as shown in Figure~\ref{fig:one-shot-mt50}.
            Co-learning is able to achieve this generalization without making strong assumptions about the structure of the observation or action space.
        }\label{fig:data-mixing}
    \end{minipage}
\end{figure}

In zero-shot and few-shot training, we train a single ``universal'' policy to perform all tasks.
We do this by training on minibatches with an equal number of (task, state, action) tuples from each task.
In the one-shot data configuration, we instead seek to train a separate model for each \textit{target task} from a single demonstration of that task by leveraging demonstrations of other \textit{base tasks}.
We achieve this by using minibatches which contain a mix of tuples sampled in equal number from all base tasks and an equal number of tuples sampled from the single target tasks demonstration, as shown in Figure~\ref{fig:data-mixing}.
This follows prior work which found such 1:1 (one-to-one) data mixing to be effective in deep Q learning methods \cite{julian2020never,lee2022spend}.

\subsection{Differentiability and Optimization}\label{differentiability}

PCBC essentially splits action selection into three steps: plan generation, query evaluation, and action decoding.
In our experiments using Language-World, query evaluation is performed with the \hyperref[query-answering-function]{query answering function}, so only the action decoder is finetuned with gradient descent.
In a real-world application of PCBC using a visual question answer (VQA) model in place of the QAF, both of the VQA model and action decoder could be tuned with gradient descent.
Our experiments also perform both query evaluation and action decoding at each timestep.
Although this already uses significantly less computational resources than evaluating a language model at each timestep, it may be possible to use fewer still computational resources by performing query evaluation at a lower frequency than every timestep.
Because the action decoder is trained as part of the end-to-end objective, it is able to compensate for minor inconsistencies in the transition between skills, such as those a lag in evaluating queries would introduce.

In this work we relied on the implicit regularization of small batch sizes (less than 200 total timesteps per minibatch) to avoid overfitting to our small dataset.
In future work, it would be worthwhile to combine our method with regularization or contrastive learning techniques that may allow improving generalization further while learning with larger batch sizes.

\section{Limitations}

\paragraph{\textbf{Reinforcement Learning}} In this work we chose to focus on using imitation learning, because PCBC required only a very small number of expert demonstrations to reach high performance.
However, significant prior research exists in using reinforcement learning (RL) for robotic control, and Meta-World (which Language-World extends), is designed as a (Meta/Multi-Task) RL benchmark.
In future work, it would be worthwhile to experiment with using plan conditioning and RL, and Offline RL in particular.

\paragraph{\textbf{Plan Quality}} In this work we use plans generated by large language models (LLMs).
Although we experimented with a variety of \hyperref[cond-plan-generation]{plan formats}, and ran each LLM multiple times for each (task, plan format) combination, many of the generated plans were still low quality.
We expect that further improvements in prompting methods, which is an area of active research, may be able to improve our \hyperref[fig:one-shot-mt50]{one-shot results} further.
Alternatively, writing more plans by hand, either to use directly or to fine-tune an LLM, may be effective.

\paragraph{\textbf{Plan Complexity}}
In this work we explored conditioning on the steps of a plan.
This is similar to evaluating a single switch statement in an end-to-end differentiable way.
In future work, we intend to explore using end-to-end differentiable models of more complex computational constructs.

\section{Conclusion}
In this work, we introduced a new benchmark, Language-World, which uses the same task as the popular Meta-World benchmark, but extends it to allow it to be easily used in experiments with large language models.
We also introduced a method, plan conditioned behavioral cloning (PCBC), which serves as a strong baseline imitation learning method for Language-World.
By leveraging Cross-Task Co-Learning, PCBC is able to achieve promising performance from a single demonstration per task, and extremely strong performance at 100 demonstrations.
In all cases, PCBC uses significantly less data than typically used by RL algorithms on these same tasks, which often require at least 10,000 episodes to achieve a non-zero success rate on a single task.

\newpage
\FloatBarrier

\end{document}